\documentclass{article}

     \usepackage[preprint]{neurips_2021}

\usepackage[utf8]{inputenc} 
\usepackage[T1]{fontenc}    
\usepackage{hyperref}       
\usepackage{url}            
\usepackage{booktabs}       
\usepackage{amsfonts}       
\usepackage{nicefrac}       
\usepackage{microtype}      
\usepackage{xcolor}         

\usepackage{amsmath}
\usepackage{amssymb}
\usepackage{enumerate}
\usepackage{subfig}
\usepackage{natbib}
\PassOptionsToPackage{numbers, square}{natbib}
\usepackage{graphicx}
\usepackage{caption}
\usepackage{float}
\usepackage{subfloat}
\usepackage{xcolor}
\usepackage{mathrsfs}
\usepackage[normalem]{ulem}
\usepackage{xcolor,cancel}
\usepackage{enumitem}
\usepackage{multirow}
\usepackage{algorithm2e}
\usepackage{arydshln}
\usepackage{verbatim}
\usepackage[toc,page,header]{appendix}
\usepackage{minitoc}

\newtheorem{theorem}{Theorem}

\DeclareMathOperator*{\argmax}{argmax}

\DeclareMathSizes{10}{9.5}{7}{5}

\title{Multi-way Clustering and Discordance Analysis through Deep Collective Matrix Tri-Factorization}

\author{%
Ragunathan Mariappan, Vaibhav Rajan\\
  Department of Information Systems and Analytics\\ School of Computing,
  National University of Singapore\\
  \texttt{\{mragunathan,vaibhav.rajan\}@nus.edu.sg} \\
}

\begin{document}

\maketitle

\begin{abstract}
Heterogeneous multi-typed, multimodal relational data is increasingly available in many domains and their exploratory analysis poses several challenges.
We advance the state-of-the-art in neural unsupervised learning to analyze such data.
We design the first neural method for collective matrix tri-factorization of arbitrary collections of matrices to perform spectral clustering of all constituent entities and learn cluster associations. 
Experiments on benchmark datasets demonstrate its efficacy over previous non-neural approaches.
Leveraging signals from multi-way clustering and collective matrix completion 
we design a unique technique, called Discordance Analysis, to reveal information discrepancies across subsets of matrices in a collection with respect to two entities. 
We illustrate its utility in quality assessment of knowledge bases and in improving representation learning. 

\end{abstract}

\doparttoc 
\faketableofcontents 

\section{Introduction}



Clustering is a fundamental technique used to discover latent group structure in data.
Multi-way or co-clustering clusters both entities -- along row and column dimensions -- of a data matrix exploiting their correlations, to discover coherent block structure within the matrix.
Multi-way clustering of a {\it collection} of matrices benefits from indirect relations during learning and is a useful tool to explore both underlying groups and their associations.
Such collections of multi-typed, multi-modal relational data, as matrices or graphs, are increasingly available in many domains, e.g., interactions of genes, diseases, drugs and clinical data in biomedicine; social network users, their posted content as images, videos, text messages, along with multiple sources of information about the content.

Multi-way clustering can be done through tri-factorization that yields, for each matrix in the collection, cluster indicators for both entities and a matrix of association values across the row and column clusters.
This approach, called CFRM, was shown to be equivalent to spectral clustering of each entity in the collection 
\citep{cfrm}.
Recent research in spectral clustering has successfully used neural representation learning in spectral clustering for single matrix inputs \citep{shaham2018spectralnet} and multi-view settings \citep{huang2019multi,yang2019deep}.
The latter assumes multiple views for the same subjects but cannot model auxiliary inputs such as data about the view features. 
These neural methods have the advantages of scalability, out-of-sample extension and ability to model complex data distributions.
This motivates the investigation of a neural method for spectral clustering in the general setting of  
{\it arbitrary} collections of matrices.
A neural extension of CFRM is challenging since it requires the design of an architecture that can jointly train for the tri-factorization and representation learning losses for any number of inputs, that may contain mixed (real and ordinal) datatypes, and ensure orthogonal spectral embeddings for each entity.
In this paper, we address these challenges by developing a
model, called Deep Collective Matrix Tri-Factorization (DCMTF) that
takes as input an arbitrary collection of matrices and learns simultaneously:
(i) real-valued latent representations of the entity instances, 
(ii) cluster structure in each entity and (iii) pairwise cluster associations based on the learnt latent representations.

In a large, heterogeneous collection of matrices, two entities may be directly and indirectly related in multiple ways, e.g., genes and diseases may have associations in a matrix and may be indirectly related through  gene--drug and drug--disease association matrices.
Thus, there may be multiple \textit{paths} through the collection connecting two entities. 
The analysis of such paths, in light of underlying clusters, in large heterogeneous collections can be a useful exploratory tool.
Consider trivial unit-length paths -- two matrices containing gene-disease associations from different sources. 
Information discrepancies between them may be discerned simply by a  cell--by--cell comparison.
However, for longer paths, i.e., indirect relations across matrices with arbitrary datatypes, there is no tool, to our knowledge, that can aid in analysis of information discrepancy across the paths.
We formalize this notion and leverage both multi-way clustering and representation learning to design a new technique to analyze multiple paths connecting entities in a collection.
We call this Discordance Analysis and illustrate its applications in quality assessment of knowledge bases and preprocessing for reprsentation learning. 

To summarize, our contributions are:
(1) We design DCMTF, the first neural method for collective matrix tri-factorization of arbitrary collections of matrices to perform spectral clustering of all constituent entities.
DCMTF is found to outperform previous non-neural tri-factorization based methods in clustering accuracy on multiple real datasets.
(2) We design a new unsupervised exploratory technique, called Discordance Analysis (DA), to analyze heterogeneous matrix collections that can reveal information discrepancies across subsets of matrices in the collection with respect to two entities. 
We illustrate two novel applications of DA:
(i) Identifying incompleteness in knowledge bases based on external data, 
(ii) Preprocessing Heterogeneous Information Networks to improve representation learning for subsequent node classification and link prediction tasks.

\section{Related Work}

Spectral clustering has been extensively studied  \citep{shi2000normalized,ng2001spectral,von2007tutorial}, in muti-view settings as well,
e.g., \citep{kumar2011co,li2015large}.
Recent neural extensions include SpectralNet \citep{shaham2018spectralnet} for a single input matrix and some methods for multi-view data \citep{huang2019multi,yang2019deep}.
None of these have been designed for arbitrary collections of matrices or multi-way clustering.
The method closest to ours is Collective Factorization of Related Matrices (CFRM) \citep{cfrm}, 
that iteratively performs eigendecomposition of a matrix similar to the Laplacian 
to obtain spectral embeddings for each entity in the input collection.

Representation learning from arbitrary collections of matrices have been studied in Collective Matrix Factorization (CMF) \citep{singh2008relational}, group-sparse CMF \citep{klami2013group} and a neural approach Deep CMF \citep{mariappan2019deep}. 
These approaches learn {\it two} latent factors for the row and column entities of each matrix, to reconstruct them. 
Another way to represent collections of relational data is through
Heterogeneous Information Networks (HIN) with multiple edge and node types
\citep{yang2020heterogeneous}.
Any HIN can be represented as a collection of matrices: each edge type represents a dyadic relation (a matrix) and each node type is an entity.
Matrices readily accommodate multiple datatypes which is not straightforward to model in HIN. Further, vectorial node features may be added as additional matrices.
Hence, a collection of matrices may be considered a more  flexible and generic model.
A tri-factorization approach for arbitrary collections of matrices, called Data Fusion by Matrix Factorization (DFMF) was designed in \citep{zitnik2015data}. 
Their goal was primarily representation learning, but since their tri-factorization is based on \citep{wang10semi}, the representations can be interpreted as soft cluster indicators if appropriate dimensions are chosen.
Associations between latent representations were learnt and used to 
analyze connections between genes and diseases in distant matrices
\citep{zitnik2016jumping}.
Unlike these methods, through simultaneous clustering and representation learning, DCMTF learns cluster-aware representations.

\section{Deep Collective Matrix Tri-Factorization (DCMTF)}
\label{sec:dcmtf}

\begin{figure*}[!h] 
\captionsetup{font=footnotesize}
\captionsetup[subfigure]{labelformat=empty}
\centering
    \subfloat[]{
    \centering
            \fontsize{5pt}{5pt}\selectfont
            \includegraphics[width=1.0\textwidth]{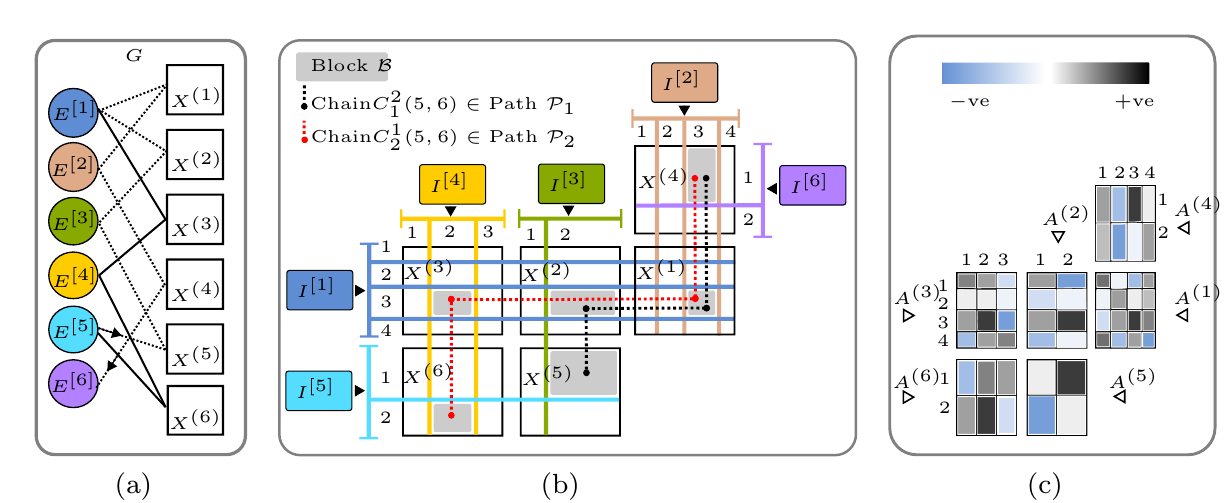}
    }
\caption{(a) Input matrices and entity-matrix graph (b) Output multi-way clusters, (c) Output cluster associations. E.g., $X^{(1)}$ has data of entities $E^{[1]}$ (rows) and $E^{[2]}$ (columns), in (a), and is tri-factorized into cluster indicators $I^{[1]}$, $I^{[2]}$ (4 clusters each), in (b), and the cluster association matrix $A^{(1)}$, in (c).
See section \ref{da:desc} for paths, chains.}
\label{fig:dcmtf:setting}
\end{figure*}

Our inputs are an arbitrary collection of $M$ matrices containing relational data among $N$ entities, 
and 
an entity-matrix 
%
bipartite relationship graph  $G(\mathcal{E},\mathcal{X},Q)$, where the vertices $\mathcal{E}$ = $\{E^{[1]},\cdots,E^{[N]}\}$ are entities, $\mathcal{X}$ = $\{X^{(1)},\cdots,X^{(M)}\}$ are matrices and edges $(E^{[e]},X^{(m)}) \in Q$ show the entities in each matrix
(fig.\ref{fig:dcmtf:setting}(a)).
Each entity has multiple {\it instances} corresponding to rows or columns in one or more matrices.
We use $e$ to index entities and $r_m,  c_m$ represent the row and column entities, respectively, in the $m^{\rm th}$ matrix ($e, r_m, c_m \in \{1,\ldots,N\}$).
We use $[e]$, $(m)$ and $(e,m)$ to denote entity $E^{[e]}$, matrix $X^{(m)}$ and edge $(E^{[e]},X^{(m)})$ respectively.
Let $d_{[e]}$ represent the number of instances of the $e^{\rm th}$ entity. 
Our aim is to collectively learn entity representations and perform multi-way clustering to obtain:
    \begin{enumerate}[noitemsep,topsep=0pt,labelindent=0em,leftmargin=*]
        \item $U^{[e]} \in \mathbb{R}^{d_{[e]} \times l}$: 
        Latent $l$-dimensional representations of all entity instances.
        \item $I^{[e]} \in \{0,1\}^{d_{[e]} \times k_{[e]}}$: Entity cluster indicators for each entity, where $k_{[e]}$ is the number of disjoint clusters of entity $E^{[e]}$ and $I_{iu}^{[e]} = 1$ indicates that the $i^{\rm th}$ instance is associated with the $u^{\rm th}$ cluster.
        \item $A^{(m)} \in \mathbb{R}^{k_{[r_m]} \times k_{[c_m]}}$:  Association matrix corresponding to each of the matrices $X^{(m)}$, containing the strength of association between clusters of the row and column entities.
    \end{enumerate}

Multi-way clustering can be done through tri-factorization: $X_{[r_m],[c_m]}^{(m)} \approx I^{[r_m]} A^{(m)} {I^{[c_m]}}^T$.
Finding such factors is NP-hard for binary indicator matrices, even for a single input matrix \citep{anagnostopoulos2008approximation}.
Spectral clustering uses a relaxation to real-valued orthonormal matrices that are obtained by solving:
$\min \sum_e Tr(C^{[e]^T} L^{[e]} C^{[e]}) \text{ subject to } C^{[e]^T}C^{[e]} = \mathbf{I}_{k_{[e]}},$
where 
$L^{[e]}$ is a Laplacian defined on a similarity metric $\mathcal{S}^{[e]}$:
$
L^{[e]} = D^{[e]} - \mathcal{S}^{[e]}; \;
D^{[e]} = \text{diag}(\sum_{\textit{row}}\mathcal{S}^{[e]} ),
$
and $\mathbf{I}_{k_{[e]}}$ is the identity matrix.
Once the spectral embeddings $C^{[e]}$ are learnt, the binary cluster indicators $I^{[e]} \in \{0,1\}^{d_{[e]} \times k_{[e]}}$ may be obtained by clustering $C^{[e]}$ using k-means. 
In CFRM, this approach is adopted with a special symmetric matrix instead of $L^{[e]}$ and the cluster associations for the $m^{\rm th}$ matrix $X^{(m)}$ is given by 
$A^{(m)} = {J}^{[r_m]^T} X^{(m)} J^{[c_m]}$, where $J^{[e]}$ is a {\it vigorous} cluster indicator, that also captures disjoint cluster memberships: $J_{iu}^{[e]} = \nu(I^{[e]}_{iu}) = I^{[e]}_{iu}/ ({|\pi^{[e]}_u|}^{\text{\tiny{1/2}}})$; $i \in \{1,\dots,d_{[e]}\}$, $u \in \{1,\dots,k_{[e]}\}$, $\pi^{[e]}_u$ is the set of $u$-th cluster entity instances and $|.|$ denotes cardinality. See \citep{cfrm} for more details. 

\subsection{Neural Collective Multi-Way Spectral Clustering Network}

\begin{figure*}[!h]
\captionsetup{font=footnotesize}
\captionsetup[subfigure]{labelformat=empty}
     \centering 
        \subfloat[]{
                \centering
                \def\svgwidth{1.0\textwidth}
                \fontsize{7pt}{7pt}\selectfont
                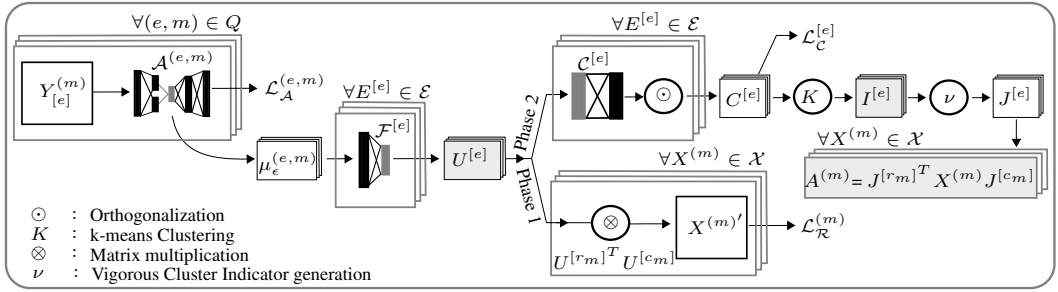 
        } 
        \caption{Schematic of DCMTF Network Construction and Training}
        \label{fig:dcmtf:model}
\end{figure*}

Our neural architecture consists of 3 subnetworks: (i) Variational autoencoders (VAE), one for each entity in each matrix, (ii) Fusion subnetworks to obtain unique entity representations and (iii) Clustering subnetworks to obtain spectral embeddings. 
The entire network (fig. \ref{fig:dcmtf:model}) is dynamically constructed as the number of VAEs and other subnetworks is determined by the number of inputs.

{\bf Entity Representations.}
Entity-specific representations may be learned by using $N$ different autoencoders, one for each entity, where each autoencoder takes as input the concatenation of all matrices containing that entity.
This approach is inadequate for matrices with different datatypes and sparsity levels as discussed in \citep{mariappan2019deep}.
Our approach addresses these problems but with higher computational cost.
Our first subnetwork, denoted by $\mathcal{A}^{(e,m)}$, has $2M$ VAEs to enable modeling of different data distributions and sparsity levels in the matrices.
Let $f_{\epsilon}, f_{\delta}$ denote the parameterized functions of the encoder and decoder respectively, realized through neural networks with weights $\epsilon,\delta$.
The encoder outputs the parameters of the learnt distribution.
Let $\mu_{\epsilon}$ denote the mean of the distribution learnt through the encoder.
For the decoder, we use Gaussian (for real-valued) or Bernoulli (for binary) outputs as described in \citep{kingma2013auto}.
Let $Y^{(m)}_{[e]}$ denote the $e^{\rm th}$ entity instances (in rows or columns) in the matrix $X^{(m)}$ along its rows.
We have $\mu_{\epsilon}^{(e,m)} = f_{\epsilon^{(e,m)}}(Y_{[e]}^{(m)}).$

{\bf Fusion.} In the second subnetwork, denoted by $\mathcal{F}^{[e]}$, these representations are fused to obtain a single real-valued representation for each entity.
If the same entity is present in more than one matrix, the final entity-specific representation is obtained by 
$U^{[e]} = f_{\eta^{[e]}} ( \Gamma_{m}
[ \mu_{\epsilon}^{(e,m)} ]  ),$
where $\Gamma_{m}[.]$ represents the concatenation operation and the index $m$ iterates over all matrices containing the $e^{\rm th}$ entity.
If the entity is present in a single matrix, no fusion is required and we set $U^{[e]} = \mu_{\epsilon}^{(e,m)}$.
The functions $f$ are realized through neural networks with weights $\eta$.

{\bf Spectral Clustering.} 
To use the learnt entity representations for clustering, we design our similarity metric defined as follows.
Let $X^{(m)^{\prime}}$ = $U^{[r_m]}U^{[c_m]^T}$ be the $m^{\rm th}$ reconstructed matrix, from the row and column entity representations learnt collectively from all inputs.
Let
$
P^{(e,m)} = X^{(m)\prime} J^{[c_m]}$ for a row entity ($e = r_m$) and  
$
P^{(e,m)} = X^{{(m)\prime}^T} J^{[r_m]}$ for a column entity ($e = c_m$), a $k_{[e]}$--dimensional representation for each entity.
For the $i^{\rm th}$ and $j^{\rm th}$ instances of entity $E^{[e]}$, we define a similarity metric using the Gaussian kernel:
$
\mathcal{S}_{P}^{[e]}(i,j)
 = \sum_{(e,m) \in Q}\text{exp}({-||P^{(e,m)}_i - P^{(e,m)}_j||^2}/{2\sigma^2}),$ 
where $\sigma > 0$ is the scale hyperparameter, 
and prove 
that 
this metric preserves the required property:
\begin{theorem}\label{thm:equiv}
The problem 
$\min \sum_{[e]} Tr(J^{[e]^T} L^{[e]} J^{[e]}) \text{ subject to } J^{[e]^T}J^{[e]} = \mathbf{I}_{k_{[e]}}; \forall E^{[e]} \in \mathcal{E}$,
is equivalent to the problem:
$
\min \sum_{(m)} (||X^{(m)} - J^{[r_m]} A^{(m)} J^{[c_m]^T}||^2)  \text{ subject to } J^{[e]^T}J^{[e]} = \mathbf{I}_{k_{[e]}} ; \forall X^{(m)} \in \mathcal{X}$
, where 
$L^{[e]}$ is a Laplacian defined on the similarity metric $\mathcal{S}_{P}^{[e]}$.
\end{theorem}

Following the usual relaxation adopted in spectral clustering, instead of $J^{[e]}$ we optimize for real-valued orthogonal $C^{[e]}$ and then use k-means on the learned spectral representations to obtain binary indicators.
We learn $C^{[e]}$ through the subnetwork denoted by $\mathcal{C}^{[e]}$,
implemented via a feedforward neural network
$f_{\gamma^{[e]}}$ with weights $\gamma$. 
To ensure orthogonal outputs $C^{[e]}$, we follow the procedure in \citep{shaham2018spectralnet,huang2019multi}.
During forward propagation through $\mathcal{C}^{[e]}$, the input to the last layer $\widetilde{C}^{[e]}$ is used to perform Cholesky decomposition:
$\widetilde{H}^{[e]}
\widetilde{H}^{{[e]}^T}
= \text{Cholesky}(
\widetilde{C}^{[e]}
\widetilde{C}^{{[e]}^T}
).$
The weight of the final linear layer is fixed to 
$\widetilde{H}^{{[e]}^{{-1}^T}}$ yielding the output
$
C^{[e]} = \widetilde{C}^{[e]} \widetilde{H}^{{[e]}^{{-1}^T}},$ 
which is orthogonal \citep{shaham2018spectralnet}.


\subsection{Network Training}




The subnetworks  $\mathcal{A}^{(e,m)}, \mathcal{F}^{[e]},\mathcal{C}^{[e]}$ are trained collectively to solve: 
$\min_{\epsilon, \delta,\gamma, \eta}
\sum_{{(e,m) \in Q}} \mathcal{L}_\mathcal{A}^{(e,m)}
+
\sum_{{e = 1}}^{N} \mathcal{L}_\mathcal{C}^{[e]} + 
\sum_{m = 1}^{M} \mathcal{L}_\mathcal{R}^{(m)}. 
$
The loss 
$
\mathcal{L}_\mathcal{A}^{(e,m)}
$
is set to be binary cross entropy for binary inputs and mean square error for real input matrices,
as described in \citep{kingma2013auto}.
The clustering loss $ \sum_{{e = 1}}^{N} \mathcal{L}_\mathcal{C}^{[e]}$ 
effects tri-factorization through trace minimization:
$\mathcal{L}_\mathcal{C}^{[e]} =  Tr\big(C^{([e])^T} L^{[e]} C^{[e]}\big).$
The matrix-specific reconstruction loss
$\sum_{m = 1}^{M} \mathcal{L}_\mathcal{R}^{(m)} $ 
enforces a factor interpretation on the latent representations used within the similarity metric:
$\mathcal{L}_\mathcal{R}^{(m)} = \ell_\mathcal{R}^{(m)} (X^{(m)}, U^{[r_m]}{U^{[c_m]}}^T),$
where $\ell_\mathcal{R}^{(m)}$ 
is set to Frobenius norm of the error for real input  and binary cross entropy for binary inputs matrices with appropriate link functions.


Optimization is done in a coordinate descent manner, using stochastic gradient descent, where we iteratively train $\mathcal{A}^{(e,m)}, \mathcal{F}^{[e]}$ (phase 1) followed by
$\mathcal{A}^{(e,m)}, \mathcal{C}^{[e]}$ (phase 2).
In phase 1, there are two steps. In the first step, we update the weights of the VAE: $\epsilon^{(e, m)}, \delta^{(e, m)}$ by backpropagating $\mathcal{L}_{\mathcal{A}}^{(e,m)}$.
In the second step, we update the weights of the fusion network and encoder: $\eta^{[e]}, \epsilon^{(e,m)}$ by backpropagating $\mathcal{L}_{\mathcal{R}}^{(m)}$.
In phase 2, there are three steps.
In the first step, we update the weights of the VAE: $\epsilon^{(e, m)}, \delta^{(e, m)}$ by backpropagating $\mathcal{L}_{\mathcal{A}}^{(e,m)}$.
Next, the weights of the last layer are computed via Cholesky as described above.
In the final step of phase 2, we update the weights of the clustering network and encoder: $\gamma^{(e)}, \epsilon^{(e,m)}$ by backpropagating $\mathcal{L}_{\mathcal{C}}^{[e]}$. Algorithm \ref{algo:dcmtf} shows a summary.

\begin{algorithm}[!h]
\captionsetup{font=footnotesize}
    \caption{Deep Collective Matrix Tri-Factorization (DCMTF)}
    \label{algo:dcmtf}
    \hrulefill
    \SetKwInOut{Input}{Inputs}
    \SetKwInOut{Output}{Outputs}
    \SetKwProg{dCMTF}{dCMTF}{}{}
    \SetKwRepeat{Repeat}{repeat}{until}

    \Input{Matrices $\mathcal{X} = {X^{(1)},...,X^{(M)}}$, Entity-matrix graph $G(\mathcal{E},\mathcal{X},Q)$,  
    \#clusters $k_{[e]}$ required
    }
    \Output{Entity representations $\mathcal{U}=U^{[1]},...,U^{[N]}$,  
    Entity cluster indicators $\mathcal{I}=I^{[1]},...,I^{[N]}$,
    Cluster associations $\mathcal{T} = {A^{{(1)}},...,A^{{(M)}}}$
    }
    \tcp{Dynamic Network Construction}
    Construct network $\mathcal{N}$ with $\mathcal{A}^{(e,m)}$; $\forall (e,m) \in Q$ and  $\mathcal{F}^{[e]}$, $\mathcal{C}^{[e]}$; $\forall E^{[e]} \in \mathcal{E}$ \\
    \tcp{Network Training}
    \Repeat{Stopping Criteria Reached}{
        $\mathcal{L}_{1}$ $\gets$ $\mathcal{L}_{\mathcal{A}}^{(e,m)}$ + $\mathcal{L}_{\mathcal{R}}^{(m)}$ $\gets$ \textit{forward}($\mathcal{A}^{(e,m)}$,$\mathcal{F}^{[e]}$) $\quad$ \tcp{phase1} 
        Backpropagate $\mathcal{L}_{1}$ and update weights:  $\epsilon^{(e, m)}, \delta^{(e, m)}$ of $\mathcal{A}^{(e,m)}$;  $\eta^{[e]}, \epsilon^{(e,m)}$ of $\mathcal{F}^{[e]}$ \\
        $\mathcal{L}_{2}$ $\gets$ $\mathcal{L}_{\mathcal{A}}^{(e,m)}$ + $\mathcal{L}_{\mathcal{C}}^{[e]}$ $\gets$ \textit{forward}($\mathcal{A}^{(e,m)}$,$\mathcal{C}^{[e]}$) $\quad$ \tcp{phase2} 
        Backpropagate $\mathcal{L}_{2}$ and update weights: $\epsilon^{(e, m)}, \delta^{(e, m)}$ of $\mathcal{A}^{(e,m)}$; $\gamma^{(i)}, \epsilon^{(e,m)}$ of
        $\mathcal{C}^{[e]}$ 
    }
    Trained $\mathcal{N}$ weights: $\epsilon_{\#}$, $\eta_{\#}$, $\gamma_{\#}$, $\delta_{\#}$ \\
    \tcp{Output generation using the trained network}
    $\mu_{\epsilon}^{(e,m)}$ = $f_{\epsilon^{(e,m)}_{\#}}(Y_{[e]}^{(m)})$; $\quad \forall (e,m) \in Q$ \tcp{Entity representation}
    %
    $U^{[e]} = \mu_{\epsilon}^{(e,m)} \text{if }e \text{ in 1 matrix, else } 
    U^{[e]} = f_{\eta^{[e]}_{\#}} ( \Gamma_{(e,m) \in Q}[ \mu_{\epsilon}^{(e,m)} ]); 
     \forall E^{[e]} \in \mathcal{E}$ \tcp{Fusion}
    %
    $C^{[e]} = f_{\gamma^{[e]}_{\#}} (U^{[e]})$;   
    $I^{[e]}$ = k-means($C^{[e]}$) with k $= k_{[e]}$; $J^{[e]}$ = $\nu$($I^{[e]}$); $\forall E^{[e]} \in \mathcal{E} $ \tcp{Clustering} 
    $A^{(m)} = J^{[r_m]^T} X^{(m)} J^{[c_m]}$; $\quad \forall X^{(m)} \in \mathcal{X}$ \tcp{Cluster associations} \hrulefill
    %
\end{algorithm}

\section{Discordance Analysis}\label{da:desc}

Collective Matrix Tri-Factorization provides three valuable intrinsic signals 
that we utilize to develop a new way of analyzing a collection of matrices.
The first signal comes from representations $U^{[e]}$ that can reconstruct the matrices: $X^{(m)^{\prime}}$ = $U^{[r_m]}U^{[c_m]^T}$. The second signal is from multi-way clustering that yields blocks $\mathcal{B}^{(m)}_{\{u,v\}}$ ($u,v$ index clusters along row and column entities, $[r_m], [c_m]$, respectively). Note that we have both reconstructed blocks, denoted by $\mathcal{B}^{(m)^\prime}_{\{u,v\}}$, and the input blocks in all matrices.

The third signal is from multiple paths in the entity-matrix graph of the collection that may connect the same two entities $E^{[i]}, E^{[j]}$.
For instance, in fig.\ref{fig:dcmtf:setting}, we see two paths between $E^{[5]}, E^{[6]}$:
one is 
$E^{[5]}$-$X^{(5)}$-
$E^{[3]}$-$X^{(2)}$-
$E^{[1]}$-$X^{(1)}$-
$E^{[2]}$-$X^{(4)}$-
$E^{[6]}$, shown in dotted lines in fig. \ref{fig:dcmtf:setting}(a) and as the black path $\mathcal{P}_1$ in fig. \ref{fig:dcmtf:setting}(b);
another is
$E^{[5]}$-$X^{(6)}$-
$E^{[4]}$-$X^{(3)}$-
$E^{[1]}$-$X^{(1)}$-
$E^{[2]}$-$X^{(4)}$-
$E^{[6]}$
shown as the red path $\mathcal{P}_2$ in fig. \ref{fig:dcmtf:setting}(b).
Note that entities $E^{[1]},E^{[2]}$ are shared across paths $\mathcal{P}_1, \mathcal{P}_2$.

After multi-way clustering, each such path across matrices may be traversed at the block level using the Association matrices.
Given cluster $u$ along the row entity $[r_m]$, the mapping $\argmax_v A^{(m)}(u,v)$ yields a (highly associated) cluster $v$ along the column entity $[c_m]$ within the $m^{\rm th}$ matrix.
Given a path $\mathcal{P}_{\rho}(i,j)$ between two entities $E^{[i]}, E^{[j]}$, for each
cluster $u \in \{1,\cdots k_{[i]}\}$ of entity $E^{[i]}$, this mapping induces a path through the blocks, that we call a {\it chain} $\mathscr{C}_{\rho}^u(i,j)$.
In fig. \ref{fig:dcmtf:setting}, entities $E^{[5]}, E^{[6]}$ connected through paths $\mathcal{P}_1, \mathcal{P}_2$ (fig. \ref{fig:dcmtf:setting}(b)) may be traversed at the block level (shown in shaded blocks), starting from a specific cluster, 
where the associations (fig. \ref{fig:dcmtf:setting}(c)) are used to select each block.

A single chain may be scored with respect to the deviation between the original and predicted blocks in the chain.
We define $D_1(\mathscr{C}_{\rho}^u(i,j)) = \sum_{\mathcal{B} \in \mathscr{C}_{\rho}^u(i,j)} d_{\text{cos}}(\mathcal{B},\mathcal{B}^{'})$, where
$d_{\text{cos}}$ is the cosine distance, averaged over the rows of the cluster blocks compared.
We compare two chains with respect to
the deviation between predicted blocks in one chain and input blocks in the other chain for the entities shared across the chains.
%
Let $\mathcal{E}_S$ denote the set of shared entities in chains $\mathscr{C}^u_w(i,j)$ and $\mathscr{C}^u_a(i,j)$ between entities $E^{[i]}, E^{[j]}$ along the paths $\mathcal{P}_w$ and $\mathcal{P}_a$ respectively.
Let blocks $\mathcal{B}^{(m_w)}_{\{n_{[s]},v\}} \in \mathscr{C}^u_w(i,j)$ and 
$\mathcal{B}^{(m_a)}_{\{n_{[s]},h\}} \in \mathscr{C}^u_a(i,j)$ belong to the shared entity $E^{[s]} \in \mathcal{E}_S$ whose row cluster index $n_{[s]}$ match, where $(m_w) \in \mathcal{P}_w$ and $(m_a) \in \mathcal{P}_a$.
These blocks may have vectors of different dimensions (from different matrix clusters ($v$ and $h$). 
So, to compute the required deviation, we use chordal distance $d_{chord}$ and define: 
$D_2(\mathscr{C}^u_w(i,j), \mathscr{C}^u_a(i,j)) = \sum_{E^{[S]} \in  \mathcal{E}_s} d_{\text{chord}}(\mathcal{B}^{(m_w)'}_{\{n_{[s]},v\}},\mathcal{B}^{(m_a)}_{\{n_{[s]},p\}})$.
Chordal distance is an extension of the Grassmann distance 
to subspaces of different dimensions and can be computed through singular value decomposition (SVD) of the product of the inputs \citep{ye2016schubert}. 

We can now analyze paths connecting two entities in a collection of matrices through the signals within and across the underlying chains.
Consider two paths $\mathcal{P}_{w}(i,j)$ and $\mathcal{P}_{a}(i,j)$ connecting entities $E^{[i]}, E^{[j]}$ through different non-overlapping subsets of matrices (there may be overlapping entities in these paths).
In each path, there are multiple chains, as many as the number of clusters in entity $E^{[i]}$.
Let $\mathscr{C}^u_w(i,j)$ and $\mathscr{C}^u_a(i,j)$ denote the chains starting from the $u^{\rm th}$ cluster of entity $E^{[i]}$ in paths $\mathcal{P}_{w}(i,j)$ and $\mathcal{P}_{a}(i,j)$ respectively.
Then, we define a score for a pair of chains as follows (we omit $(i,j)$ in all chains to reduce clutter):
$
S_u(\mathscr{C}^u_w, \mathscr{C}^u_a) = 
\alpha D_1(\mathscr{C}^u_w) - \beta D_1(\mathscr{C}^u_a) - \gamma D_2(\mathscr{C}^u_w, \mathscr{C}^u_a),
$
where $\alpha, \beta, \gamma$ are the weights for each of the distances. We define a {\it discordant} pair of chains in the two paths $\mathcal{P}_{w}(i,j)$ and $\mathcal{P}_{a}(i,j)$ as the one with the highest score $S_u$.

The intuition behind the scoring is as follows.
Suppose the collection has matrices from two independent sources, one more reliable than the other, e.g., from expert-curated knowledge bases and observations/operational data.
Let us call these Knowledge and Data matrices.
Consider path $\mathcal{P}_{w}(i,j)$ connecting entities $E^{[i]}, E^{[j]}$ through Knowledge matrices and path $\mathcal{P}_{a}(i,j)$
connecting the same two entities through the Data matrices respectively.
Then, a discordant pair of chains indicates a chain of cluster blocks in the knowledge matrices that differ in their information content from the chain in the data matrices across the same two entities.
Note that such discordance may be computed for paths across any two non-overlapping subsets of matrices in the collection.
The signal for this discordance is derived from cluster-aware shared latent representations, learnt collectively, and their predictions.

Each term in $S_u$ plays a role in analysing this discordance.
High $D_1(\mathscr{C}^u_w)$ directly signals high deviation between known and predicted Knowledge.
All the entity representations are collectively learnt for both Knowledge and Data matrices.
So, if the representations that yield low $D_1(\mathscr{C}^u_w)$ also yield low distance $D_1(\mathscr{C}^u_a)$ between observed and predicted {\it Data}, then it indicates a concordance across the chains, inferred from the effect of the learnt representations.
Similarly, high $D_1(\mathscr{C}^u_w)$ and high $D_1(\mathscr{C}^u_a)$ also indicate concordance.
Chain pairs with high $D_1(\mathscr{C}^u_w)$ and low $D_1(\mathscr{C}^u_a)$ indicates higher discordance -- hence, the negative weight for the second term.
Further, we consider the entity cluster blocks shared across paths $\mathcal{P}_{w}$ and $\mathcal{P}_{a}$ in $D_2$. 
For these entity representations in different matrices of the chain, a low distance $D_2$, between the predicted Knowledge and input Data, is preferred for high discordance through the negative weight.
The goal is to identify chains where despite low $D_2$ of shared representations in these shared blocks, there is discordance observed through $D_1(\mathscr{C}^u_w)$ and $D_1(\mathscr{C}^u_a)$ which gives a stronger signal of information discrepancy across the paths.

\section{Experimental Results}
\label{sec:expt}

\subsection{Clustering Performance}

We compare the clustering performance of DCMTF with that of CFRM \citep{cfrm} and DFMF \citep{zitnik2015data}. 
Standard clustering metrics are used: Rand Index (RI), Normalized Mutual Information (NMI) along with chance-adjusted metrics Adjusted Rand Index (ARI) and Adjusted Mutual Information (AMI).
We use four real datasets that have 3,5,7 and 10 matrices as shown in Table \ref{tab:DCMTF}. 
A single entity is chosen in each collection for clustering.
Most matrices are more than 99\% sparse, with more than 1000 instances in most entities. 
Wiki and PubMed data are discussed further in the following sections.
DCMTF outperforms both the methods in all the metrics, except for RI on 
Freebase,
indicating that neural representations learn the cluster structure better.

\begin{table}[!h]
    \centering
    \captionsetup{font=footnotesize}
    \footnotesize
    \begin{tabular}{c|p{1.65em}p{1.65em}p{2.75em}|p{1.65em}p{1.65em}p{2.75em}|p{1.65em}p{1.65em}p{2.75em}|p{1.65em}p{1.65em}p{2.75em}}
    \toprule
     \textbf{Dataset:} & \multicolumn{3}{|c}{\textbf{Wiki}} & \multicolumn{3}{|c}{\textbf{Cancer}} & \multicolumn{3}{|c}{\textbf{Freebase}} & \multicolumn{3}{|c}{\textbf{PubMed}} \\ 
     & \multicolumn{3}{|c}{$M=3, k=3$} & \multicolumn{3}{|c}{$M=5, k=4$} & \multicolumn{3}{|c}{$M=7, k=8$} & \multicolumn{3}{|c}{$M=10, k=8$} \\ 
     \hline
     \textbf{Metric} & DFMF & CFRM & DCMTF & DFMF & CFRM & DCMTF & DFMF & CFRM & DCMTF & DFMF & CFRM & DCMTF \\ \hline
    ARI &   0.0533  &   0.0103  &   \textbf{0.4189}  &   0.0045  &   0.0094  &   \textbf{0.0147}  &   0.0420  &   0.0419  &   \textbf{0.0660}  &   0.0060  &   0.0014  &   \textbf{0.0200}  \\
    AMI &   0.1147  &   0.0156  &   \textbf{0.3789}  &   0.0051  &   0.0081  &   \textbf{0.0173}  &   0.0694  &   0.0509  &   \textbf{0.1473}  &   0.0055 & $\text{-}$0.0033 &   \textbf{0.0183}  \\
    NMI &   0.1284  &   0.0247  &   \textbf{0.3796}  &   0.0105  &   0.0139  &   \textbf{0.0220}  &   0.0830  &   0.0805  &   \textbf{0.1595}  &   0.0481  &   0.0451  &   \textbf{0.0516}  \\ 
    RI  &   0.5290  &   0.4347  &   \textbf{0.7354}  &   0.5410  &   0.5422  &  \textbf{0.5464}  &   \textbf{0.7723}  &   0.6234  &   0.6608  &   0.4976  &   0.3680  &   \textbf{0.6334}  \\
     \bottomrule
    \end{tabular}
    \caption{Clustering Performance. $M$: Number of matrices in collection, $k$: \#clusters for a chosen entity (input).}
    \label{tab:DCMTF}
\end{table}

\subsection{Case Study: Synchronizing Wikipedia Infoboxes through Discordance Analysis}

A Wikipedia article has 2 parts: (i) a body of unstructured information on the article's subject, and (ii) an {\it infobox}, a fixed-format table that summarizes key facts from the article.
Wikidata is a secondary database that collects structured data to support Wikipedia.
It extracts and transforms different parts of Wikipedia articles to create a relational knowledge base, that can be programmatically accessed.
Wikidata can be used to automatically create and maintain Wikipedia's infoboxes across different languages
\citep{vrandevcic2014wikidata}.
An ongoing challenge for Wikidata is to maintain its completeness \citep{balaraman2018recoin,wisesa2019wikidata}, in real-time and at the scale of up to 500 edits/minute, across more than 300 languages.
Completeness refers to the degree to which information is not missing \citep{zaveri2016quality}, e.g., with respect to attributes in an Infobox.


We illustrate the utility of Discordance Analysis (DA) as a tool to identify blocks of information (as cluster chains) that are likely to contain  discordance with respect to completeness. 
We selected Wikidata items 
from 3 categories -- movies, books and games, which form the clusters to which the items belong. 
Each item has only one category in our dataset.
These items are associated with one or more subjects, e.g., the movie item ``Terminator 3: Rise of the Machines'' is associated with the subjects `android' and `time travel'.
Both the items and subjects have Wikipedia entries accessible via Wikidata webservices. We obtain the abstract section of these articles and create bag-of-words vector representations from them for each item and subject.
Our dataset, thus, comprises 3 entities -- items $[t]$, subjects $[z]$ and bag-of-words vectors $[b]$ in 3 matrices $X_{([t],[z])}^{(1)}$, $X_{([b],[z])}^{(2)}$, $X_{([b],[t])}^{(3)}$ as shown in fig. \ref{fig:wiki}. 
The matrix $X_{([t],[z])}^{(1)}$ is binary indicating item-subject associations, while the other two have counts.
There are 1800 items, 1116 subjects and the vocabulary size is 3883.

\begin{minipage}{\textwidth}
\captionsetup{font=footnotesize}
  \begin{minipage}[b]{0.44\textwidth}
    \centering
    \centering
            \def\svgwidth{0.9\textwidth} 
            \fontsize{8pt}{8pt}\selectfont
            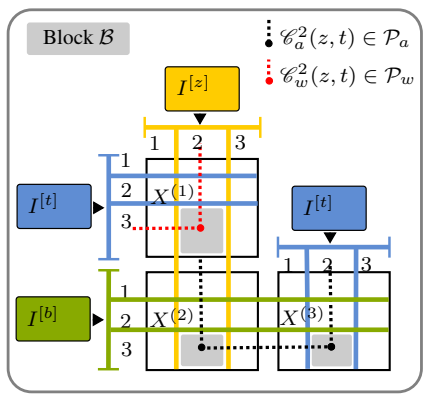
    \captionof{figure}{3 entities $[t]$: Items, $[z]$: Subjects, $[b]$: Bag-of-Words vectors and 3 relations between the entities -- matrices $X_{([t],[z])}^{(1)}$, $X_{([b],[z])}^{(2)}$, $X_{([b],[t])}^{(3)}$.}
    \label{fig:wiki}
  \end{minipage}
  \hfill
  \begin{minipage}[b]{0.53\textwidth}
    \centering
    \footnotesize
\begin{tabular}{ccccc}
\toprule
\textbf{Method} & \textbf{Entity}
& \multicolumn{3}{c}{\textbf{$\mathscr{C}$\# and \% of entities found}}  \\ \hline
&& \textit{$\mathscr{C}$1}           & \textit{$\mathscr{C}$2}     & \textit{\textbf{*}$\mathscr{C}$3} \\ \cline{3-5}
\multirow{3}{*}{{DCMTF + DA}}
& [b] &   31.50    &   16.67   &   51.83   \\
& [z] &   18.75   &   12.50    &   68.75   \\
& [t] &   0.00   &   0.00   &   \textbf{100.00} \\\cdashline{2-5}
& total &   50.25   &   29.17   &   \textbf{220.58}  \\

\hline
&& \textit{\textbf{*}$\mathscr{C}$1}           & \textit{$\mathscr{C}$2}     & \textit{$\mathscr{C}$3}\\ \cline{3-5}
\multirow{3}{*}{{DFMF + DA}}
& [b] &   17.40   &   17.40   &   23.99   \\
& [z] &   37.50   &   37.50   &   25.00   \\
& [t] &   \textbf{63.64}   &   0.00    &   36.36   \\ \cdashline{2-5}
& total&   \textbf{118.54}  &   54.90   &   85.35   \\
\hline
&& \textit{$\mathscr{C}$1}           & \textit{\textbf{*}$\mathscr{C}$2}     & \textit{$\mathscr{C}$3}\\ \cline{3-5}
\multirow{3}{*}{{CFRM + DA}}
& [b] &   51.65   &   30.40   &   17.95   \\
& [z] &   18.75   &   18.75   &   6.25    \\
& [t] &   0.00    &   \textbf{90.91}   &   9.09    \\ \cdashline{2-5}
& total &   70.40   &   \textbf{140.06}  &   33.29   \\
\bottomrule
\end{tabular}
\captionof{table}{\textbf{*}: most discordant cluster chain selected by DA for each Method; \textbf{Bold:} \% item/total entity instances found in the selected cluster chains.}
\label{tab:wiki}
    \end{minipage}
  \end{minipage}

We randomly select and hide 5\% of the items and their subjects to which they belong in $X_{([t],[z])}^{(1)}$.
The experiment is repeated for 4 such random selections.
The corresponding bag-of-words vectors are also considered hidden.
If this association is missing in Wikidata ($X_{([t],[z])}^{(1)}$) but the Wikipedia entries for the item and subject contain enough signal (in our case, $X_{([b],[z])}^{(2)}$, $X_{([b],[t])}^{(3)}$) to associate the two, then we expect DA to find it within a discordant cluster chain.
We consider $X_{([t],[z])}^{(1)}$ as the \textit{Knowledge} matrix and the matrices $X_{([b],[z])}^{(2)}$, $X_{([b],[t])}^{(3)}$ as \textit{Data}.
As shown in fig. \ref{fig:wiki}, there are two paths
$\mathcal{P}_w, \mathcal{P}_a$ (red and black respectively) connecting the entities $[t]$ and $[z]$.
Multi-way clustering is done using DCMTF, CFRM and DFMF, with k=3 in all cases, corresponding to 3 item categories.
These clustering results (for items) are shown in Table \ref{tab:DCMTF}.
Considering 3 chains along $\mathcal{P}_w$ and 3 chains along $\mathcal{P}_a$ (along each cluster), there are 9 pairs that are scored for DA and the highest is chosen.

\textbf{Results.}
Table \ref{tab:wiki} shows the percentage of hidden entities found in the cluster chains selected by DA for one random selection. 
With all three methods, DA is able to find the cluster chain with highest discordance (or maximum percentage of hidden items) in items.
Among the chosen chains, the one obtained via DCMTF has the highest total percentage and the maximum number of hidden items.
This is expected as simultaneous clustering and representation learning in DCMTF supports DA.
These results demonstrate the ability of DA to detect information discrepancy across chains. We illustrate DA on a 
setting of 3 matrices, but it can be used with any number of 
matrices and is also expected to detect other discrepancies with respect to quality, e.g., consistency or timeliness.



\subsection{Case Study: Improving Network Representation Learning using Discordance Analysis}

We illustrate an application of DA in data cleaning for representation learning from heterogeneous information networks (HIN).
Since HIN can be represented as a collection of matrices, DCMTF can be used for multi-way clustering of a HIN and a cluster chain represents a subset of edges within the HIN.
A discordant cluster chain is a subset of edges that is intrinsically discordant with respect to the edges in the HIN, across different edge types, and the links they can predict.
Hence, we hypothesize that removing discordant cluster chains may serve as a useful preprocessing or cleaning step that can improve learning of node representations (or embeddings) from the `cleaned' HIN.

We consider two popular random-walk based HIN embedding methods Metapath2vec \citep{dong2017metapath2vec} and HIN2Vec \citep{fu2017hin2vec}, which have been extended for HIN from Deepwalk \citep{perozzi2014deepwalk}. 
The key idea is to generate random walks on the graph and use a neural network to predict the neighborhood in the walks of a node, essentially adapting skip-gram architectures \citep{mikolov2013distributed} for graphs.
In a HIN, the random walks are guided
by user-specified metapaths that provide a semantically meaningful scheme for the walks. 
For instance, in a citation network, with node types Authors $[a]$ and Papers $[r]$, edge types `cited by', `written by', metapaths could be 
$[r]$-$[r]$, $[r]$-$[a]$, $[r]$-$[r]$-$[r]$, $[r]$-$[r]$-$[a]$ etc., to constrain the walks.

To perform DA on HIN, we pick an entity of interest and the direct and indirect relations present in the HIN for the entity.
For example, in the citation network, $[r]$-$[r]$ is a direct relation, while $[r]$-$[a]$ and $[a]$-$[r]$ form an indirect relation between $[r]$ and $[r]$. 
Similarly, $[r]$-$[a]$ is a direct relation and $[r]$-$[r]$-$[a]$ and $[a]$-$[r]$-$[r]$  form indirect relations between $[r]$ and $[a]$.
A direct relation connects the entities through a single matrix while indirect relation connects the same two entities through multiple matrices.
For DA, we assume matrices in direct relations as knowledge matrices and matrices in indirect relations as data matrices and perform the analysis.
For each direct relation, we consider all the indirect relations to find discordant chains. 
We remove the union of edges found over all discordant edges, for all the direct relations with respect to an entity, 
to `clean' the HIN.

We used the PubMed dataset provided as one of the benchmark datasets in \citep{yang2020heterogeneous} to evaluate HIN Embedding techniques. 
The dataset consists of 10 matrices and 4 entities (fig. \ref{fig:hin:pubmed}). 
We selected a random subset of entity instances as listed in Table \ref{tab:pubmed} to train the embeddings after holding out 10\% of the disease-disease edges (for link prediction).
Following \citep{yang2020heterogeneous}, embeddings are evaluated on 2 tasks: node classification and link prediction.
For node classification, 8 disease labels for about 10\% of the diseases are given, disease embeddings are used as feature vectors, and metrics used are macro-F1 (across labels) and micro-F1 (across nodes).
For link prediction, the Hadamard function is used to construct feature vectors for node pairs, and metrics used are AUC and MRR.
A linear SVC classifier is used and 5-fold CV is done for both tasks.
We learn the embeddings in 3 different ways:

\begin{enumerate}[noitemsep,topsep=0pt,labelindent=0em,leftmargin=*]
\item
Original HIN.
We use the data 
directly to learn node embeddings using HIN2Vec and metapath2vec. 
For metapath inputs in these methods, we use all the paths by which the four entities Gene $[g]$, Disease $[d]$, Chemical $[c]$ and Species $[p]$ are directly or indirectly related to the entity Disease as seen in fig. \ref{fig:hin:pubmed}. 
These include (i) direct path $[c]$-$[d]$ and indirect paths
$[c]$-$[g]$-$[d]$, $[c]$-$[p]$-$[g]$-$[d]$, $[c]$-$[p]$-$[d]$
and (ii) direct path $[p]$-$[d]$ and indirect path $[p]$-$[g]$-$[g]$-$[d]$.
\item 
DA-Cleaned.
We consider matrices with direct paths as knowledge and those with indirect paths as data, we perform DA with (i) entities $[c]$ and $[d]$ and (ii) entities $[p]$ and $[d]$.
We use DCMTF with $k=8$ for DA.
The clustering results (for diseases) are shown in Table \ref{tab:DCMTF}.
All the edges in the discordant cluster chains found, 1028 edges in total, are removed.
We run HIN2Vec and Metapath2vec on this cleaned HIN.

\item
Rand-Cleaned.
As a baseline for comparison we randomly remove edges from the 10 matrices.
We iterate through the matrices and randomly remove a subset of edges until the total number of edges removed crosses 1028.
Then, we run HIN2Vec and Metapath2vec 
to obtain embeddings. 
\end{enumerate}

\begin{minipage}{\textwidth}
\captionsetup{font=footnotesize}
    \begin{minipage}[b]{0.35\textwidth}
        \captionsetup{font=footnotesize}
        \label{fig:fig:hin:sample}
        \centering
            \def\svgwidth{0.84\textwidth} 
            \fontsize{10pt}{10pt}\selectfont
\begingroup%
  \makeatletter%
  \providecommand\color[2][]{%
    \errmessage{(Inkscape) Color is used for the text in Inkscape, but the package 'color.sty' is not loaded}%
    \renewcommand\color[2][]{}%
  }%
  \providecommand\transparent[1]{%
    \errmessage{(Inkscape) Transparency is used (non-zero) for the text in Inkscape, but the package 'transparent.sty' is not loaded}%
    \renewcommand\transparent[1]{}%
  }%
  \providecommand\rotatebox[2]{#2}%
  \newcommand*\fsize{\dimexpr\f@size pt\relax}%
  \newcommand*\lineheight[1]{\fontsize{\fsize}{#1\fsize}\selectfont}%
  \ifx\svgwidth\undefined%
    \setlength{\unitlength}{102.21836481bp}%
    \ifx\svgscale\undefined%
      \relax%
    \else%
      \setlength{\unitlength}{\unitlength * \real{\svgscale}}%
    \fi%
  \else%
    \setlength{\unitlength}{\svgwidth}%
  \fi%
  \global\let\svgwidth\undefined%
  \global\let\svgscale\undefined%
  \makeatother%
  \begin{picture}(1,0.85510545)%
    \lineheight{1}%
    \setlength\tabcolsep{0pt}%
    \put(0,0){\includegraphics[width=\unitlength,page=1]{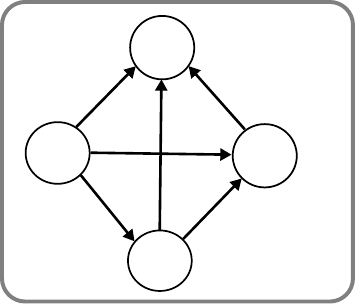}}%
    \put(0.42029535,0.75028738){\color[rgb]{0,0,0}\makebox(0,0)[lt]{\begin{minipage}{0.9000485\unitlength}\raggedright $[d]$\end{minipage}}}%
    \put(0.11085588,0.45011251){\color[rgb]{0,0,0}\makebox(0,0)[lt]{\begin{minipage}{0.9000485\unitlength}\raggedright $[c]$\end{minipage}}}%
    \put(0.41473656,0.14808473){\color[rgb]{0,0,0}\makebox(0,0)[lt]{\begin{minipage}{0.9000485\unitlength}\raggedright $[p]$\end{minipage}}}%
    \put(0.69482714,0.44766368){\color[rgb]{0,0,0}\makebox(0,0)[lt]{\begin{minipage}{0.9000485\unitlength}\raggedright $[g]$\end{minipage}}}%
    \put(0,0){\includegraphics[width=\unitlength,page=2]{pubmed_hin.pdf}}%
    \put(0.87200549,0.65032366){\color[rgb]{0,0,0}\makebox(0,0)[lt]{\begin{minipage}{1.21803133\unitlength}\raggedright $\mathcal{P}_a$\end{minipage}}}%
    \put(0,0){\includegraphics[width=\unitlength,page=3]{pubmed_hin.pdf}}%
    \put(0.86994864,0.77172209){\color[rgb]{0,0,0}\makebox(0,0)[lt]{\begin{minipage}{1.21803133\unitlength}\raggedright $\mathcal{P}_w$\end{minipage}}}%
  \end{picture}%
\endgroup%

        \captionof{figure}{Pubmed HIN; Nodetypes: Gene $[g]$, Disease $[d]$, Chemical $[c]$, Species $[p]$. Edgetypes in Table \ref{tab:pubmed}.}
        \label{fig:hin:pubmed}
    \end{minipage}
  \hfill
  \begin{minipage}[b]{0.65\textwidth}
    \centering
    \footnotesize
        \begin{tabular}{p{2em}p{3em}p{2.75em}p{2.75em}p{1.5em}p{1.5em}p{1.75em}p{1em}}
            \toprule
            \textbf{Matrix} & \textbf{Relation (Edge)} & \textbf{Row Entity} & \textbf{Col Entity} & \textbf{Row Dim} & \textbf{Col Dim} & \textbf{\#edges} & \textbf{DA}\\ \hline
            $X^{(1)}$ & associates & Gene & Gene & 2634 & 2634 & 5142 & 0.1 \\ 
            $X^{(2)}$ & causing & Gene & Disease & 2634 & 4405 & 4851 & 2.8 \\ 
            $X^{(3)}$ & associates & Disease & Disease & 4405 & 4405 & 5732 & 0.1 \\ 
            $X^{(4)}$ & in & Chemical & Gene & 5660 & 2634  & 5727 & 9.8 \\ 
            $X^{(5)}$ & in & Chemical & Disease & 5660 & 4405 & 7494 & 4.2 \\ 
            $X^{(6)}$ & associates & Chemical & Chemical & 5660 & 5660 & 8747 & 0.0 \\ 
            $X^{(7)}$ & in & Chemical & Species & 5660 & 608 & 910 & 0.0\\ 
            $X^{(8)}$ & with & Species & Gene & 608 & 2634 & 639 & 0.3 \\ 
            $X^{(9)}$ & with & Species & Disease & 608 & 4405 & 810 & 0.5 \\ 
            $X^{(10)}$ & associates & Species & Species & 608 & 608 & 137 & 0.0 \\ \bottomrule
    \end{tabular}
    \captionof{table}{PubMed 
    (Dim: Dimension, DA: \%edges removed by DA)}
    \label{tab:pubmed}
    \end{minipage}
  \end{minipage}

\textbf{Results.} Table \ref{tab:hne} shows the results on the two benchmark tasks using embeddings from the data directly (Original HIN) and after cleaning. 
We observe that random removal of edges does not improve link prediction AUC and MRR with Metapath2vec while the improvement with HIN2Vec is not significant. Node classification improves marginally with both methods, when edges are randomly removed.
With edges removed through DA, the performance improvement for both tasks across all the metrics is significant (p-value < 0.0038, via Friedman test \citep{demvsar2006statistical}).
These results support our hypothesis that discordant chains in the HIN may obfuscate representation learning. Our cleaning can also be seen as a DA-based subsampling procedure informed by DCMTF's clusters and representations.

\begin{table}[!h]
    \centering
    \captionsetup{font=footnotesize}
    \footnotesize
    \begin{tabular}{ccccccccc}
    \toprule
          & \multicolumn{4}{c}{HIN2Vec} & \multicolumn{4}{c}{Metapath2Vec} \\ \hline
         \textbf{Network} & \multicolumn{2}{c}{\textbf{Node Classification}} & \multicolumn{2}{c}{\textbf{Link Prediction}} & \multicolumn{2}{c}{\textbf{Node Classification}} & \multicolumn{2}{c}{\textbf{Link Prediction}} \\ 
          & Macro-F1 & Micro-F1 & AUC & MRR & Macro-F1 & Micro-F1 & AUC & MRR \\ \hline
    Original HIN & 0.0891 & 0.1390 & 0.7952 & 0.9502 & 0.1085 & 0.1259 & 0.6573 & 0.8299\\     
    Rand-Cleaned & 0.1029 & 0.1522 & 0.7999 & 0.9487 & 0.119 & 0.1392 & 0.6539 & 0.8196\\
    DA-Cleaned & \textbf{0.1224} & \textbf{0.1655} & \textbf{0.8156} & \textbf{0.9525} & \textbf{0.1380} & \textbf{0.1620} & \textbf{0.6664} & \textbf{0.8543} \\
    \bottomrule
    \end{tabular}
    \caption{Mean performance scores over 5-fold CV in disease node classification and disease-disease link prediction using embeddings learnt, through HIN2Vec and Metapath2Vec, from HIN and its 2 `cleaned' versions.}
    \label{tab:hne}
\end{table}

\section{Conclusion}
\label{sec:concl}
We advance the state-of-the-art for unsupervised analysis of heterogeneous collections of data through the design of two neural methods.
We improve multi-way spectral clustering in arbitrary collections of matrices by developing DCMTF, the first neural method for this task,
that empirically outperforms previous non-neural methods.
Further, we utilize multi-way clustering and collective matrix completion within cluster chains 
to design a fundamentally new unsupervised exploratory technique, called Discordance Analysis (DA), 
that can reveal information discrepancies with respect to two entities.
We illustrate two novel applications of DA in quality assessment of knowledge bases and preprocessing HIN for representation learning. 
Future work may investigate ways to improve the computational complexity of DCMTF. Additional uses of DA, e.g., in information visualization, and its information-theoretic characterization would also be interesting to explore.



\small

\bibliographystyle{unsrtnat}
\bibliography{main.bib}

\end{document}